%% file: main.tex
\documentclass[10pt,twocolumn,letterpaper]{article}

\usepackage{cvpr}
\usepackage{times}
\usepackage{epsfig}
\usepackage{graphicx}
\usepackage{amsmath}
\usepackage{amssymb}
\usepackage{animate}
\usepackage{comment}

\usepackage[pagebackref=true,breaklinks=true,letterpaper=true,colorlinks,bookmarks=false]{hyperref}

\input{config/02_pkgs.tex}
\input{config/00_literals.tex}

\input{config/01_macros.tex}

\def\paperID{105} 
\def\confName{3DV\xspace}
\def\confYear{2024\xspace}

\newcommand\extrafootertext[1]{%
    \noindent
    \bgroup
    \renewcommand\thefootnote{\fnsymbol{footnote}}%
    \renewcommand\thempfootnote{\fnsymbol{mpfootnote}}%
    \footnotetext[0]{#1}%
    \egroup
}

\begin{document}

\title{\vspace{-5mm}\acronym: Human and Camera Motion Estimation from in-the-wild Videos}

\author{Muhammed Kocabas\textsuperscript{1,2,3} \quad 
        Ye Yuan\textsuperscript{1} \quad
        Pavlo Molchanov\textsuperscript{1}  \quad
        Yunrong Guo\textsuperscript{1} \quad 
        Michael J. Black\textsuperscript{2} \quad  \\
        Otmar Hilliges\textsuperscript{3} \quad 
        Jan Kautz\textsuperscript{1} \quad 
        Umar Iqbal\textsuperscript{1} \\
        \small{
\textsuperscript{1}NVIDIA 
\qquad  
\textsuperscript{2}
Max Planck Institute for Intelligent Systems, Tubingen, Germany 
\qquad \textsuperscript{3}ETH Zurich, Switzerland
}
\\
\url{https://nvlabs.github.io/PACE/}
}

\input{figures/teaser}
\extrafootertext{\hspace{-5mm} The work was done during Muhammed's internship at NVIDIA.}

\input{sections/00_abstract.tex}

\input{sections/01_intro.tex}

\input{sections/02_rwork.tex}

\input{sections/03_method.tex}

\input{sections/04_results.tex}

\input{sections/05_conclusion.tex}

\appendix
\input{sections/06_appendix.tex}

{\small
\bibliographystyle{ieee_fullname}
\bibliography{references}
}

\end{document}

%% file: config/02_pkgs.tex
\usepackage{graphicx}
\usepackage{tikz}
\usepackage[accsupp]{axessibility}  
\usepackage{enumitem} 
\usepackage{comment} 
\usepackage{multirow}
\usepackage{pifont}
\usepackage{float}
\usepackage{enumitem}
\setitemize{noitemsep,topsep=0pt,parsep=0pt,partopsep=0pt}
\usepackage{balance}
\usepackage{dsfont}
\usepackage{lipsum}  
\usepackage[thinc]{esdiff}

\usepackage{times}
\usepackage{epsfig}
\usepackage{graphicx}
\usepackage{float}
\usepackage{amsmath}
\usepackage{amssymb}
\usepackage{soul}
\usepackage{multicol}
\usepackage{blindtext}
\usepackage{here}
\usepackage{caption}
\usepackage{makecell}
\usepackage[normalem]{ulem}
\usepackage{multirow}
\usepackage{booktabs}
\usepackage{xcolor, colortbl}
\usepackage{xspace}
\usepackage{balance}
\usepackage{numprint}
\usepackage{sidecap}
\usepackage[accsupp]{axessibility}  

%% file: config/00_literals.tex
\newcommand{\acronym}{PACE\xspace}
\newcommand{\droid}{DROID-SLAM\xspace}
\newcommand{\hybrik}{HybrIK\xspace}

\newcommand{\HCM}{HCM\xspace}

\newcommand{\worldErr}{W-MPJPE\xspace}
\newcommand{\worldAlignedErr}{WA-MPJPE\xspace}
\newcommand{\worldRootErr}{W-RJE\xspace}
\newcommand{\paErr}{PA-MPJPE\xspace}
\newcommand{\accel}{ACCEL\xspace}
\newcommand{\ate}{ATE\xspace}
\newcommand{\ateWoS}{ATE-S\xspace}
\newcommand{\camAccel}{CAM ACCEL\xspace}



%% file: config/01_macros.tex
\definecolor{mypurple}{RGB}{0, 128, 255}
\definecolor{dexcolor}{RGB}{0, 117, 4}
\definecolor{honotatecolor}{RGB}{1, 0, 253}

\excludecomment{comments}

\excludecomment{remove}

\newcommand{\PM}[1]{{\color{blue}[PM: #1]}}
\newcommand{\UI}[1]{{\color{magenta}[UI: #1]}}

\definecolor{GreenColor}{rgb}{0.137,0.573,0.565}
\definecolor{OrangeColor}{rgb}{0.914,0.541,0.0.141}
\definecolor{PurpleColor}{rgb}{0.5,0,0.7}
\definecolor{BlueColor}{rgb}{0,0.725,0.949}
\definecolor{PinkColor}{rgb}{0.9843,0.19215,0.6}



%% file: figures/teaser.tex
\twocolumn[{
	\renewcommand\twocolumn[1][]{#1}
	\maketitle
	\begin{center}
		\newcommand{\teaserwidth}{\textwidth}
		\vspace{-0.15in}
		\centerline{
			\includegraphics[width=\teaserwidth,clip]{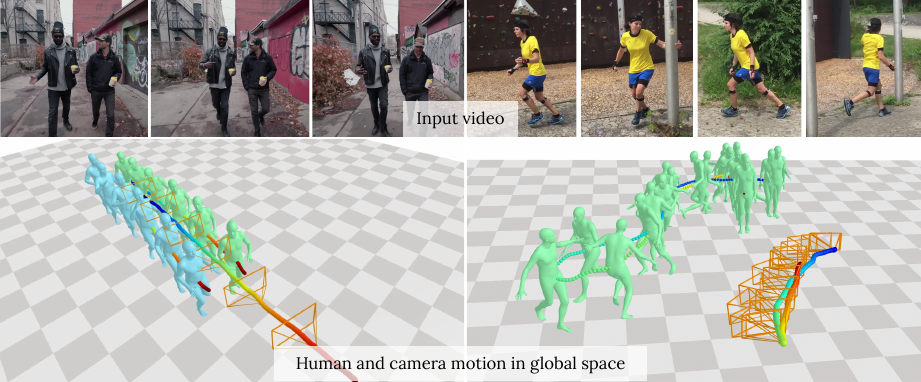}
		}
		\vspace{-1ex}
		\captionof{figure}{\textbf{Human and camera motion reconstruction from in-the-wild videos:} given a video of multiple people, \acronym is able to reconstruct the motions of all humans and the camera in a \textit{coherent} global space. To achieve this, we leverage the benefits of both camera localization methods and human motion priors, exploiting the complementary nature of these approaches, \ie, dynamic foreground motion \vs static background features, to address each other's limitations.}
		\vspace{-0.08in}
		\label{fig:teaser}
	\end{center}
 }]

%% file: sections/00_abstract.tex
\begin{abstract}
We present a method to estimate human motion in a global scene from moving cameras. This is a highly challenging task due to the coupling of human and camera motions in the video. To address this problem, we propose a joint optimization framework that disentangles human and camera motions using both foreground human motion priors and background scene features. Unlike existing methods that use SLAM as initialization, we propose to tightly integrate SLAM and human motion priors in an optimization that is inspired by bundle adjustment. Specifically, we optimize human and camera motions to match both the observed human pose and scene features. This design combines the strengths of SLAM and motion priors, which leads to significant improvements in human and camera motion estimation. We additionally introduce a motion prior that is suitable for batch optimization, making our approach significantly more efficient than existing approaches. Finally, we propose a novel synthetic dataset that enables evaluating camera motion in addition to human motion from dynamic videos. Experiments on the synthetic and real-world RICH datasets demonstrate that our approach substantially outperforms prior art in recovering both human and camera motions.
\end{abstract}

%% file: sections/01_intro.tex
\vspace{-3mm}
\section{Introduction}

Jointly estimating global human and camera motion from dynamic RGB videos is an important problem with numerous applications in areas such as robotics, sports and mixed reality. However, it is a very challenging task because the observed human and camera motions in the video are entangled. 
Estimating human motion by itself from videos is highly under-constrained since subject and camera motion are interchangeable. Analogously, camera motion estimation is more challenging in dynamic scenes due to spurious correspondences. Finally, pure monocular approaches can only estimate camera trajectories up to scale.

There are only a few works that address the problem of global pose estimation~\cite{li2022d,yuan2022glamr,yi2022human}. 
These methods leverage the insight that the global human root trajectory is correlated with the local body movements;  \eg, observing a running motion is indicative of forward motion. 
Hence, they suggest that global root trajectories can be estimated by exploiting learned motion priors~\cite{yuan2022glamr} or by enforcing physics-based constraints on the reconstructed human motion~\cite{yi2022human, li2022d}. 
While this idea can help to estimate global human trajectories, motion priors or physical constraints are not enough to fully resolve the ambiguity in the mapping from local motion to global trajectories, especially under root rotations. Others utilize SLAM  methods (\eg, COLMAP) to estimate camera poses~\cite{liu20204d, ye2023slahmr}, then keep the camera poses fixed and estimate the global scale. However, in-the-wild videos often contain moving objects which can degrade the camera pose localization and subsequently affect the human motion estimates. 

In this paper, we propose a novel approach, called \acronym (Person And Camera Estimation), to tackle the above problems. 
We formulate the problem as a global optimization and jointly optimize human and camera motions, leveraging a bundle adjustment objective to match both human pose and background scene features. In this way, the SLAM algorithm uses mostly static scene features, that do not correspond to human motion. Simultaneously, the human motion prior helps correct inaccurate camera trajectories that are incompatible with the local body movements, and informs about the global scale based on human motion statistics. 
We show that this formulation provides robustness to inaccurate initial human or camera motion estimates. 

A further contribution lies in the human motion prior itself. Commonly used human priors \eg,~HuMoR~\cite{rempe2021humor} are typically autoregressive and become prohibitively slow when incorporated in a per-frame optimization, in particular for long motion sequences. 
In this work, we show that neural motion field (NeMF~\cite{he2022nemf}) can be used to design a parallel motion prior that drastically improves computational efficiency.
We divide the entire sequence into overlapping clips and maximize the likelihood of the human motion under the prior. This results in a significantly more efficient implementation without compromising reconstruction quality. Notably, the parallel motion prior allows the runtime of PACE to grow sub-linearly \wrt the sequence length in contrast to the linear rate in prior work.

Since it is difficult to obtain ground-truth human and
camera poses for in-the-wild videos, we also propose a new synthetic
dataset for benchmarking human and camera motion
estimation from dynamic videos called the Human and Camera Motion (HCM) dataset. It is the first
dataset that provides ground-truth human and camera motion
information for this task. We will make the dataset
publicly available to facilitate research in this direction.

We evaluate \acronym on two datasets: the newly proposed synthetic \HCM dataset and the RICH dataset~\cite{huang2022cap}, which contains a moving camera with ground truth 3D human pose and shape. Results show that our method substantially outperforms state-of-the-art (SOTA) approaches in accurately recovering human motions from dynamic cameras. Notably, our method also significantly improves camera motion estimation over SOTA SLAM algorithms for this task, which demonstrates the advantage of our global optimization framework. Additionally, we conduct extensive ablation studies to validate the impact of various design choices on performance.

In summary, our contributions are as follows:
\vspace{1mm}
\begin{itemize}
\itemsep1mm 
    \item We present a novel approach for precise global human and camera motion estimation from dynamic cameras, which tightly integrates human motion priors and SLAM into a unified optimization framework that leverages both human pose and scene information.
    \item We propose a parallel motion prior optimization scheme, which significantly improves efficiency without sacrificing accuracy, and allows the runtime to grow sub-linearly \wrt the sequence length.
    \item We introduce \HCM, a  
    synthetic dataset for benchmarking global human and camera motion estimation.
    \item Our method outperforms the SOTA methods significantly in recovering both human and camera motions, achieving 52\% and 74\% improvements respectively, which fully demonstrate the synergy of our unified approach.
\end{itemize}

%% file: sections/02_rwork.tex
\section{Related Work}
\label{sec:related_work}
\noindent\textbf{Camera-Space Human Pose Estimation.}
Due to the difficulty in monocular depth estimation, most existing methods estimate human poses in the coordinate frame centered around the pelvis of the human body~\cite{Akhter:CVPR:2015, bogo2016keep, lassner2017unite, hmrKanazawa18, pavlakos2018humanshape, guler2019holo, kolotouros2019spin, pavlakos2019texture, Rong_2019_ICCV, kolotouros2019convolutional, choutas2020expose, zanfir2020weakly, sun2019human, joo2021eft, choi2020pose, kundu2020mesh, SMPL-X:2019, xu2019denserac, monototalcapture2019, song2020human, zhang2020object, zhou2021monocular, moon2020i2l, lin2021end, Mueller:CVPR:21, kolotouros2021prohmr, Zhang_2021_ICCV, Sun_2021_ICCV, humanMotionKanazawa19, kocabas2020vibe, luo20203d, choi2020beyond, rempe2021humor}. These methods adopt an orthographic camera projection model and ignore the absolute 3D translation of the person with respect to the camera. To overcome this limitation, recent methods estimate human meshes in the camera coordinates~\cite{zanfir2018monocular, jiang2020coherent, Zanfir_2021_ICCV, ICG, Zhang_2021_CVPR, Xie_2021_ICCV, PhysCapTOG2020, liu20204d, li2020hybrik, iqbal2021kama, reddy2021tesstrack}. Some methods use an optimization framework to recover the absolute translation of the person~\cite{mono20173dhp, mehta2017vnect, XNect_SIGGRAPH2020,zanfir2018deep, rogez2017lcr} or exploit various scene constraints to improve depth prediction~\cite{zanfir2018monocular, Weng_2021_CVPR}. Others employ physics-based constraints to ensure the physical plausibility of the estimated poses~\cite{PhysCapTOG2020, Xie_2021_ICCV, GraviCap2021, yuan2021simpoe, isogawa2020optical, gartner2022trajectory}, use limb-length constraints~\cite{iqbal2020learning} or approximate depth using the bounding box size~\cite{jiang2020coherent, moon2019camera, Zhang_2021_CVPR}. Several approaches employ inverse kinematics to estimate human meshes with absolute translations in the camera coordinates~\cite{li2020hybrik, iqbal2021kama}. Heatmap-based representations have also been used to directly predict the absolute depths of multiple people~\cite{Fabbri_2020_CVPR,zhen2020smap,Sun:CVPR:2022}. A few methods learn to also predict the camera parameters from the image, which are used for absolute pose regression in the camera coordinates~\cite{Kocabas_SPEC_2021, Zanfir_2021_ICCV, li2022cliff}. While these methods achieve impressive results for camera-relative pose estimation, they fail to decouple human and camera motions from dynamic videos, and therefore cannot recover global human trajectories as our method does.

\vspace{1mm}
\noindent\textbf{Global Human Pose Estimation.}
The majority of current methods for estimating 3D poses in world coordinates rely on synchronized, calibrated, and static multi-view capture setups~\cite{belagiannis20143d, joo2018total, reddy2021tesstrack, multiviewpose, zhang20204d, dong2021shape, zhang2021lightweight, zheng2021deepmulticap, huang2021dynamic, dong2021fastpami, dai2022hsc4d, dai2023sloper4d}. Huang~\etal~\cite{wang2021dynamic} use uncalibrated cameras but still assume time synchronization and static camera setups. Hasler~\etal~\cite{hasler2009markerless} handle unsynchronized moving cameras but assume multi-view input and rely on an audio stream for synchronization. Recently, Dong~\etal~\cite{dong2020motion} proposed to recover 3D poses from unaligned internet videos of different actors performing the same activity from unknown cameras, assuming that multiple viewpoints of the same pose are available in the videos. Luvizon~\etal~\cite{luvizon2023scene} estimate the global human poses of multiple people using the scene point cloud for static cameras. In contrast, our approach estimates human meshes in global coordinates from \emph{monocular} videos recorded with dynamic cameras. Several methods rely on additional IMU sensors or pre-scanned environments to recover global human motions~\cite{vonMarcard2018, hps2021Vladmir, pavlakos2022one}, which is impractical for large-scale adoption. Another line of work has recently focused on estimating accurate human-scene interaction~\cite{hassan2019resolving, luo2021dynamics, yi2022human, huang2022cap}. Recent work uses human motion priors~\cite{yuan2022glamr} and physics-based constraints~\cite{li2022d,yi2022human} to decouple human and camera motions but does not consider background scene features, which limits performance on in-the-wild videos. Liu~\etal~\cite{liu20204d} obtain global human pose using SLAM and convert the pose from the camera to global coordinates. BodySLAM~\cite{henning2022bodyslam} uses features of both humans and scenes, but it only demonstrates results of a single unoccluded person slowly walking in an indoor scene. Along this line, a recent work~\cite{ye2023slahmr} obtains initial camera trajectories with SLAM and optimizes the scale of the camera trajectories using a human motion prior~\cite{rempe2021humor}. In contrast, our approach tightly integrates SLAM and human motion priors into a joint optimization framework, where the entire SLAM camera trajectories (not only scale) are optimized jointly to match observed human pose and background scene features. This not only leads to more accurate human trajectory estimation but also improves full camera trajectory estimation over SLAM significantly, which has not been achieved by prior work. Additionally, our parallel motion optimization scheme also makes our approach substantially (50 times) faster than~\cite{ye2023slahmr} for a sequence of 1000 frames. Our parallel scheme also allows PACE's time cost to grow sub-linearly \wrt sequence length in contrast to the linear rate of~\cite{ye2023slahmr}.

\vspace{1mm}
\noindent\textbf{Human Motion Prior.}
There has been a significant amount of research on 3D human dynamics for various tasks, including motion prediction and synthesis~\cite{fragkiadaki2015recurrent,jain2016structural,li2017auto,martinez2017human,villegas2017learning,pavllo2018quaternet,aksan2019structured,gopalakrishnan2019neural,yan2018mt,barsoum2018hp,yuan2019diverse,yuan2020dlow,yuan2020residual,cao2020long,petrovich2021action,hassan2021stochastic}. Recently, human pose estimation methods have started to incorporate learned human motion priors to help resolve pose ambiguity~\cite{kocabas2020vibe,rempe2021humor,zhang2021learning}. Motion-infilling approaches have also been proposed to generate complete motions from partially observed motions~\cite{hernandez2019human,kaufmann2020convolutional,harvey2020robust,khurana2021detecting}. Diffusion models~\cite{sohl2015deep} have also been used as priors for motion synthesis and infilling~\cite{tevet2022human,zhang2022motiondiffuse,yuan2022physdiff,huang2023diffusion}. Recently, He~\etal~\cite{he2022nemf} proposes the neural motion field (NeMF), which expresses human motion as a time-conditioned continuous function and demonstrates superior motion synthesis performance. Our approach extends NeMF by leveraging it as a motion prior for human pose estimation. Additionally, our proposed parallel motion optimization scheme enables efficient optimization of human motions.

%% file: sections/03_method.tex
\section{Method}
\input{figures/method_overview}
The input to \acronym is an in-the-wild RGB video $\mathbf{I}{=}\{\mathbf{I}_{1},\cdots,\mathbf{I}_T$\} with $T$ frames captured by a moving camera.  Our goal is to estimate both the camera motion and the motion of all visible people in the video in a global world coordinate system. The camera motion $\{\mathbf{R}_t, \mathbf{T}_t\}_{t=1}^T$ consists of the camera rotation $\mathbf{R}_t \in \mathbb{R}^{3\times3}$ and translation $\mathbf{T}_t \in \mathbb{R}^3$ for every timestep $t$ in the video. The global motion $\mathbf{Q}^i{=}\{Q^i_t{=}\{\Phi^i_t, \tau^i_t, \theta^i_t, \beta^i\}\}_{t=s^i}^{e^i}$ for person $i$ consists of the global translation $\tau^i_t \in \mathbb{R}^{3}$, global orientation $\Phi^i_t \in \mathbb{R}^{3\times3}$, and the body pose parameters $\theta^i_t \in \mathbb{R}^{23\times3}$ for all time steps $t\in \{s^i \cdots e^i\}$, where $s^i$ and $e^i$ correspond to the first and last frame in which person $i$ is visible. 
The body shape parameters $\beta^i$ are shared across all time steps.  We use the SMPL body model~\cite{SMPL:2015} to obtain the articulated body meshes $\mathbf{V}^i{=}\{V^i_t\}_{t=s^i}^{e^i}$ from $\mathbf{Q}^i$. Specifically, SMPL consists of a linear function $\mathcal{M}(\Phi, \tau, \theta, \beta)$ that maps the body motion $Q^i_t{=}(\Phi^i_t, \tau^i_t, \theta^i_t, \beta^i_t)$ to a triangulated body mesh $V^i_t \in \mathbb{R}^{6890\times3}$ with $6890$ vertices. In the rest of this paper, we drop the superscript $i$ from all variables for brevity but always assume the visibility of multiple people.

Our key insight is to harness the complementary properties of SLAM and human motion priors. The human motion prior can be used to explain foreground human motion, which typically is dynamic and therefore has been treated as unwanted noise in existing SLAM algorithms. Leveraging the motion prior in a joint optimization regularizes the camera trajectories to be in agreement with plausible human motion and provides information about the global scale. On the other hand, SLAM leverages mostly static background features, which provide information about the camera motion and can be leveraged to resolve ambiguity in the motion space of the human motion priors.

We introduce a novel unified framework, illustrated in  Fig.~\ref{fig:overview}, that simultaneously recovers the camera and human motion using a joint optimization objective (Sec.~\ref{sec:global_opt}). Since this is a highly ill-posed problem, we exploit data-driven models to initialize our objective (Sec.~\ref{sec:init}) and use human motion priors to constrain the solution space (Sec.~\ref{sec:motion_prior}).  

\subsection{Initialization}
\label{sec:init}
 We start by obtaining bounding box sequences for all visible subjects using an off-the-shelf multi-object tracking and re-identification algorithm~\cite{zhang2022bytetrack}. We then estimate body pose information for each detected bounding box using the state-of-the-art method \hybrik~\cite{li2020hybrik}. \hybrik provides body poses in the camera coordinate frame which we represent as $\hat{Q}^c_t{=}(\hat{\Phi}^c_t, \hat{\tau}^c_t, \hat{\theta}_t, \hat{\beta}_t)$. The super-script $c$ corresponds to the camera coordinate frame. Note that the local body pose $\theta_t$ and shape $\beta_t$ are agnostic to camera motion. For videos recorded with dynamic cameras, the estimated translation $\hat{\tau}^c_t$ and root orientation $\hat{\Phi}^c_t$ must be transformed from camera coordinates to a consistent world coordinate frame. This requires knowledge of the per-frame camera-to-world transforms $\{R_t,T_t\}_{t=1}^T$. For this, we leverage a data-driven SLAM method, namely \droid~\cite{teed2021droid},
which uses the information of the static scene to estimate per-frame camera-to-world transforms $\{\hat{R}_t, \hat{T}_t\}_{t=1}^T$. SLAM methods, however, provide camera translations $\hat{T}_t$ up to scale. Hence, at this stage, we only use the camera rotation information to obtain a person's root orientation in the world coordinate frame: $\hat\Phi_t = \hat{R} ^{-1}_{t} \hat{\Phi}^c_t. $
We then use a neural network similar to~\cite{li2021task,he2022nemf}
to estimate the initial global root translations $\{\hat{\tau}_t\}_{t=s}^e$ from the local pose parameters $\{\hat{\Phi}_t, \hat{\theta}_t\}_{t=s}^e$.  We use a single value for shape parameters $\beta$ for each person that we initialize with the average of the per-frame estimates from \hybrik i.e., $\hat{\beta}{=}\frac{\sum_{t=s}^e \hat{\beta}_t}{e - s}$. This forms our initial estimate of the global human motion  $\hat{Q}{=}\{\hat{Q}_t{=}(\hat{\Phi}_t, \hat{\tau}_t, \hat{\theta}_t, \hat{\beta})\}_{t=s}^e$ in the world coordinate frame. In the remainder of this paper, our goal is to refine these initial estimates via human motion priors and the background scene features, while recovering accurate global camera trajectories.

\subsection{Human Motion Prior}
\label{sec:motion_prior}
Our goal is to develop a human motion prior that ensures that the estimated human motion is plausible and also helps constrain the solution space during joint optimization of human and camera motion. For this, we use a variational autoencoder (VAE)~\cite{kingma2013auto}, which learns a latent representation $\mathbf{z}$ of human motion  
and regularizes the distribution of the latent code to be a normal distribution.  We want the decoder $\mathcal{D}$ of the VAE to be non-autoregressive for faster sampling while not sacrificing accuracy. This is important because we want to use the motion prior in an iterative optimization, and auto-regressive motion priors (e.g., HuMoR~\cite{rempe2021humor}) are prohibitively slow when processing large motion sequences. In contrast, a non-autoregressive decoder can be evaluated for the entire sequence in parallel. To this end, we adopt a Neural Motion Field (NeMF)~\cite{he2022nemf} based decoder to represent body motion
as a continuous vector field of body poses via a NeRF-style MLP~\cite{mildenhall2020nerf}. In Sec.~\ref{sec:global_opt}, we show that NeMF can be extended to a parallel motion prior that enables efficient optimization.
We follow~\cite{he2022nemf} and only model the local body motion via the prior. 
Specifically, $\mathcal{D}$ is an MLP that takes the latent codes $\{\mathbf{z}_\Phi, \mathbf{z}_\theta\}$ and a time step $t$ as input and produces the orientation $\hat{\Phi}_t$, local body pose $\hat{\theta}_t$, and joint contacts $\hat{\kappa}_t$ for a given time step:
\begin{equation}
    \mathcal{D} : (t, \mathbf{z}_\Phi, \mathbf{z_\theta}) \rightarrow (\hat{\Phi}_t, \hat{\theta}_t, \hat{\kappa}_t), 
\end{equation}
where $\mathbf{z}_\Phi$ and $\mathbf{z}_\theta$ control the root orientation $\Phi$ and the local body pose $\theta$ of the person, respectively. For a given pair of $\mathbf{z}_\Phi$ and $\mathbf{z}_\theta$ the entire sequence can be sampled in parallel by simply varying the values of $t$. To incorporate the motion priors during global optimization, we optimize the latent codes $\{\mathbf{z}_\Phi, \mathbf{z_\theta}\}$ instead of directly optimizing the local body motion $\{\Phi_t, \theta_t\}_{t=s}^e$. We initialize the latent codes using the pre-trained encoders of the VAE; \ie, $z_\Phi{=}\mathcal{E}_\Phi(\{\Phi\}_{t=s}^e$) and $z_\theta{=}\mathcal{E}_\theta(\{\theta\}_{t=s}^e$). We refer to~\cite{he2022nemf} for training details. 

\vspace{2mm}
\noindent\textbf{Global Translation Estimation.}~%
We use a fully convolutional network to generate the global translation $\tau^i_t$ of the root joint, based on the local joint positions, velocities, rotations, and angular velocities as inputs. All quantities can be computed from joint rotations. 
Our approach, which is similar to \cite{Zhou2020GenerativeTL,Li2021TaskGenericHH}, takes into account the fact that the subject's global translation is conditioned on its local poses. In order to avoid any ambiguity in the output, we predict the velocity $\dot{\tau_t}$ rather than $\tau_t$ directly, and then integrate the velocity using the forward Euler method to obtain $\tau_{t+1}{=}\tau_t + \dot{\tau_t} \Delta{t}$. We also predict the height of the root joint using the same convolutional network to prevent any cumulative errors that could cause the subject to float above or sink into the ground. 

Since changing the latent codes $\{\mathbf{z}_\Phi, \mathbf{z}_\theta\}$ also impacts the global translations $\tau_t$, for simplicity, we refer to the mapping from latent codes to global human motion as
\begin{equation}
\mathcal{P}: (t, \mathbf{z}_\Phi, \mathbf{z}_\theta) \rightarrow (\hat{\Phi}_t, \hat{\theta}_t, \hat{\tau}_t). 
\end{equation}

\label{sec:global_hpe} 

\vspace{-2mm}
\subsection{Global Optimization}

\label{sec:global_opt}
Here we detail the proposed optimization formulation for the joint reconstruction of global human and camera motion. Our goal is to optimize the latent code $\mathbf{z}{=}\{\mathbf{z_\Phi}$, $\mathbf{z_\theta}\}$ and camera-to-world transforms $\{R_t, sT_t\}$ with correct scale $s$. Note that SLAM methods assume the camera at the first frame ($t=0$) to be at the origin. To align all coordinate frames, we also optimize the camera height $h_0$ and orientation $R_0$ for the first frame.  More specifically, we optimize the following objective function:

\begin{equation} \label{eq:final_objective}
\begin{split}
    \min_{ \substack{\beta, \mathbf{z} \\ s , h_0, R_0, \{R_t, T_t\}_{t=1}^T}}  
    E_\textrm{body} + E_\textrm{scene} + E_\textrm{camera},
\end{split} 
\end{equation}
where
\begin{align*} 
    E_\textrm{body} &=  
    E_\textrm{2D} + E_{\beta} + E_{\textrm{pose}} + E^\textrm{b}_\textrm{smooth} \\ 
    & \quad\quad\quad + E_\textrm{VAE} + E_\textrm{consist}, \\
    E_\textrm{scene} &= E_\textrm{contact} + E_\textrm{height}, \\
      E_\textrm{camera} &= E_\textrm{PCL} + E^ \textrm{c}_\textrm{smooth}. 
\end{align*}

The error term $E_\textrm{body}$ ensures that the reconstructed human motion is plausible and agrees with the image evidence. $E_\textrm{2D}$ measures the 2D reprojection error between the estimated 3D motion and 2D body joints $\mathbf{x}_t$ obtained using a state-of-the-art 2D joint detector~\cite{xu2022vitpose}: 
\begin{equation} \label{eq:reprojection}
    E_\textrm{2D} = \sum_{i=1}^N \sum_{t=s_i}^{e_i} \omega_t\zeta(\Pi(R_0R_tJ^i_t + sT_t+   \begin{bmatrix} 0 \\ 0 \\ h_0  \end{bmatrix}) ~-~\mathbf{x}^i_t). 
\end{equation}
 Here $\omega_t$ are the body joint detection confidences, $\zeta$ is the robust Geman-McClure function~\cite{geman_statistical_1982}, $\Pi$ corresponds to perspective projection using the known camera intrinsic matrix $K$, and $J^i_t$ corresponds to 3D body joints that are obtained from the SMPL body mesh via a pre-trained regressor $\mathcal{W}$: 
\begin{equation}
    J^i_t = \mathcal{W}(\mathcal{M}(\mathcal{P}(\mathbf{z},t), \beta^i_t)).
\end{equation}
The error term $E_\textrm{pose}$ penalizes large deviations of the local body pose $\hat{\theta}_t$ from the \hybrik predictions,
$E_\beta$ is prior over body shapes~\cite{hmrKanazawa18},
and $E_\textrm{VAE}$ a motion prior loss defined as:
\begin{equation}
\begin{split}
    \label{eqn:vae}
    E_\textrm{VAE} = -\sum_i^N
    & \log \mathcal{N}(\mathbf{z}_{\Phi}^i; \mu_\Phi(\{\Phi^i_t\}),\sigma_\Phi(\{\Phi^i_t\}))~+ \\  
    & \log \mathcal{N}(\mathbf{z}_{\theta}^i; \mu_\theta(\{\theta^i_t\}),\sigma_\theta(\{\theta^i_t\})).
\end{split}
\end{equation}
The term $E_\textrm{contact}$ encourages zero velocities for joints that are predicted to be in contact $\hat{\kappa}_t$ with the ground plane:
\begin{equation}
E_\textrm{contact} = \sum_{i=1}^N \sum_{t{=}s_i}^{e^i} \hat{\kappa}_t^{i} || J^i_t - J^i_{t-1} ||^2,
\end{equation}
where $\hat{\kappa}_t^{i} \in \mathbb{R}^{24}$ is the contact probability output from the motion prior decoder  $\mathcal{D}$ for each joint.  $E_\textrm{height}$ prevents in-contact joints from being far away from the ground plane: 
\begin{equation}
E_\textrm{height} = \hat{\kappa}_t^{i} \textrm{max}(|J^i_t|-\delta, 0).   
\end{equation}

The ground plane is kept fixed and assumed to be $xy$-plane aligned with $+z\textrm{-axis}$ as the up direction. This parameterization allows us to optimize all variables in this consistent coordinate frame without the need to optimize an additional ground plane equation.

The error term $E_\textrm{camera}$ in Eq.~\eqref{eq:final_objective} ensures that the reconstructed camera motion is smooth and consistent with the static scene motion.
Since \droid is trained on videos with static scenes only, its estimates can be noisy due to the dynamic humans present in our target videos. Hence, we propose to use the point cloud recovered by SLAM as a direct constraint in our optimization, instead of directly relying on the camera predictions. To ensure that the points on dynamic humans do not influence camera reconstruction, we remove all points that lie inside the person bounding boxes. The term $E_\textrm{PCL}$ then computes the re-projection error of the pruned point cloud similar to Eq.~\eqref{eq:reprojection}.  The term $E^\mathrm{b}_\textrm{smooth}$ ensures that the optimized parameters are temporally smooth.

We empirically chose the weights of different error terms in our objective and provide more details in the appendix (Table~\ref{tab:opt_stages_detailed}). 
\vspace{1mm}

\noindent\textbf{Parallel Motion Optimization.} Our specific choice of human motion prior, NeMF~\cite{he2022nemf}, allows us to design a parallel motion prior that is suitable for batch optimization, which significantly enhances the efficiency of our approach.
Concretely, we split a motion sequence into overlapping windows of $T{=}128$ frames. We use 16 overlapping frames to help reduce jitter and discontinuities across windows. Dividing motions into overlapping windows also allows the latent codes of the prior to model a fixed length of motion. 
Since our motion prior is non-autoregressive, we can optimize all windows in parallel. 
To ensure smooth transitions between clips we additionally compute a batch consistency term $E_\textrm{consist}$, defined as the $\ell_2$
distance between 3D joints $J^i_t$ of overlapping frames.

\vspace{1mm}
\noindent\textbf{Multi-Stage Optimization.}%
\input{tables/opt_stages}
The task of reasoning about the camera and human motion from a video is inherently ill-posed, as optimizing both camera motion $R_t. T_t$ and motion prior latent codes $\{\mathbf{z}_{\Phi}, \mathbf{z}_{\theta}\}$ simultaneously can result in local minima. To address this challenge, we adopt a multi-stage optimization pipeline, with different parameters optimized in different stages to avoid bad minima. After obtaining initial camera motion results from SLAM and human motion results from the motion prior, the optimization process is carried out in four stages, as outlined in Table~\ref{tab:opt_stages}. In Stage-1, we optimize only the first frame camera parameters $(R_0, h_0)$, camera scale $s$, and the subjects' body shape $\beta$ based on the initial camera and human motion. 
In Stage-2, we incorporate the global orientation latent code $\mathbf{z}_\Phi$ to jointly adjust the subjects' global orientation and camera. In Stage-3, we optimize the local body motion $\mathbf{z}_\theta$ as well. Finally, in Stage-4, we jointly optimize the full camera trajectory along with $\mathbf{z}_\Phi$ and $\mathbf{z}_\theta$. Each stage is run for 500 steps. The $\lambda$ coefficients used for each objective term can be found in the appendix (Table~\ref{tab:opt_stages_detailed}). 

\vspace{1mm}
\noindent\textbf{Occlussion Handling.} Our approach offers a natural solution for occlusions due to subjects in the scene. 
We achieve this by excluding error terms for occluded frames during optimization and solely optimize the latent codes $\{\mathbf{z}_\Phi,\mathbf{z}_\theta\}$ for visible frames. After optimization, we sample motions from the motion prior to infill the missing poses which will be consistent with their visible neighbors.

%% file: figures/method_overview.tex
\begin{figure*}[t]
    \centering
    \includegraphics[width=\linewidth]{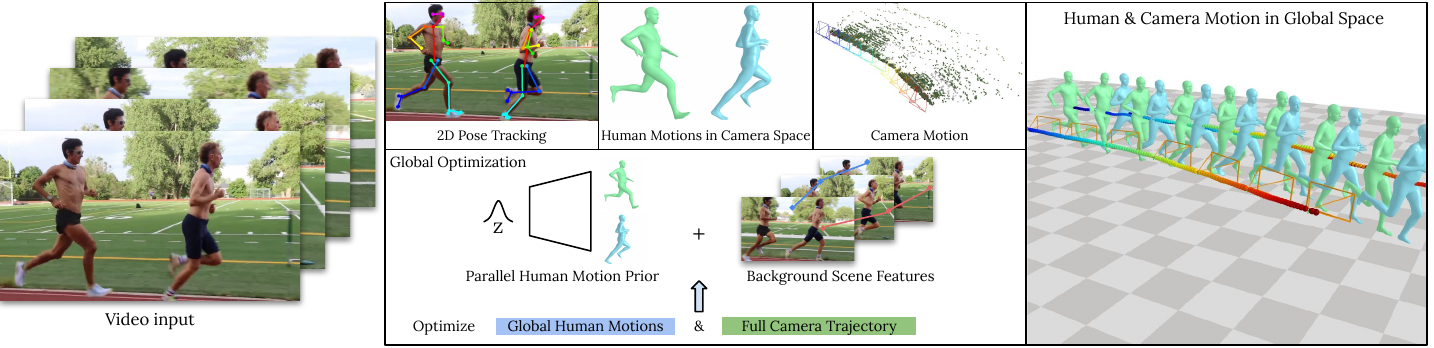}
    \caption{\textbf{\acronym overview.} Given a video with dynamic human and camera motions, we first use off-the-shelf methods to obtain initial 2D human pose, 3D human motion, and camera motions. We propose a unified optimization framework that optimizes the global human motions and full camera trajectories to reduce 2D pose errors, increase motion likelihood under human motion prior, and match background features. The final output is coherent human and camera motion in global space.} 
    \label{fig:overview}
    \vspace{-2ex}
\end{figure*}{}

%% file: tables/opt_stages.tex
\begin{table}
    \centering
    \resizebox{\linewidth}{!}{

        \begin{tabular}{l|c|c|l}
        \toprule
        \textbf{Stages} & \textbf{Opt. Variables} & \textbf{Loss Functions} & \textbf{Description}\\
        \midrule 
        Stage-1 & $s, h_0, R_0, \beta$ & $E_{2D}+E_\beta$ & camera traj. transform \\
        \midrule 
        Stage-2 & $s, h_0, R_0, \beta, \mathbf{z}_\Phi$ & $E_\textrm{body}+E_\textrm{scene}$ & + global human orientation\\
        \midrule 
        Stage-3 & $s, h_0, R_0, \beta, \mathbf{z}_\Phi, \mathbf{z}_\phi$ & $E_\textrm{body}+E_\textrm{scene}$ & + local body pose\\
        \midrule 
        Stage-4 & $\beta, \mathbf{z}_\Phi, \mathbf{z}_\phi, R_t, T_t$ & $E_\textrm{body}+E_\textrm{scene}+E_\textrm{camera}$ & + full camera trajectory\\
        \bottomrule
        \end{tabular} 
    }
    \vspace{-3mm}
   \caption{Optimization stages.}
   \vspace{-6mm}
    \label{tab:opt_stages}
\end{table}

%% file: sections/04_results.tex
\section{Experiments}

\label{sec:exp}
We design our experiments to answer the following questions: (1) Can our unified approach, PACE, achieve SOTA human motion estimation performance for dynamic videos? (2) Can PACE improve camera motion estimation of a SOTA SLAM method? (3) What are the critical components in PACE that significantly impact performance?
\input{tables/hcm_sota_slahmr}
\input{tables/rich_sota_slahmr}

\subsection{Datasets and Metrics}

\input{figures/hcm_dataset}

\noindent\textbf{HCM Synthetic Dataset.}
Currently, available datasets that provide dynamic videos  (\eg,~\cite{vonMarcard2018,liu20204d}) for evaluating human pose and shape estimation have been primarily focused on evaluating the accuracy of local body estimation while neglecting the importance of global human motion estimation. Furthermore, evaluation datasets for simultaneous localization and mapping (SLAM) algorithms do not feature humans and do not provide human motion information. As such, there is a need to create a comprehensive dataset that provides accurate labels for global human and camera motion. To address this need, we have created the HCM (Human and Camera Motion) dataset, which enables the evaluation of both human and camera motion. %
We use the characters from the RenderPeople~\cite{renderpeople} dataset and animate them in the scenes obtained from Unreal Engine marketplace~\cite{unreal}. %
We obtain motion capture (MoCap) clips from the AMASS dataset~\cite{AMASS:ICCV:2019}. For camera trajectory, we designed heuristics to replicate typical camera movements observed in everyday videos and professional movies. Final images were rendered using NVIDIA Omniverse. Additional information regarding the data generation process can be found in the appendix (Sec.~\ref{sec:hcm_dataset}). Some example sequences can be seen in Fig~\ref{fig:hcm_dataset}. 

\vspace{1mm}
\noindent\textbf{RICH Dataset.}
The RICH dataset \cite{huang2022cap} was collected using a total of 7 static and one moving camera. While the ground truth poses are available for the persons and static cameras, the ground truth poses of the moving camera are not available. As such, we only assess the performance of global human motion estimation using this dataset.

\vspace{1mm}
\noindent\textbf{Metrics.}
We report various metrics for both human and camera motion, with an emphasis on those that compute the error in world coordinates. Regarding human motion evaluation, the \worldErr metric is used to report MPJPE after aligning the first frames of the predicted and ground truth data. The \worldAlignedErr metric is used to report MPJPE after aligning the entire trajectories of the predicted and ground truth data using Procrustes Alignment. Additionally, the \paErr metric is employed to report the MPJPE error after aligning every frame of the predicted and ground truth data. We also include an \accel metric that measures the joint acceleration difference between ground-truth and predicted human motion. For camera motion evaluation, we follow SLAM methods and report the average translation error (\ate)  after rigidly aligning the camera trajectories, the average translation error without scale alignment (\ateWoS), and the \camAccel camera acceleration error. The \ateWoS metric provides a more accurate reflection of inaccuracies in the captured scale of the scene. 









\input{figures/qualitative}

\subsection{Comparison with State-of-the-Art Methods}

\noindent\textbf{Human Motion Estimation.} We compare PACE with the following baselines on the HCM and RICH datasets: GLAMR~\cite{yuan2022glamr}, SLAHMR~\cite{ye2023slahmr}, SOTA global human and camera estimation approaches; HybrIK~\cite{li2020hybrik} + SLAM, which estimates the camera motions using \droid~\cite{teed2021droid} and then transforms the human motion estimated by HybrIK from camera to world space. As observed in Tables~\ref{tab:hcm_ablation} and \ref{tab:rich_sota}, PACE outperforms GLAMR, SLAHMR and HybrIK in human motion estimation significantly. In particular, PACE drastically reduces the global pose errors, \ie, decreasing W-MPJPE by 24\% and WA-MPJPE 27\% on the HCM dataset, and reducing W-MPJPE by 40\% and WA-MPJPE by 52\% on the RICH dataset. PACE can also recover accurate local human pose, as indicated by better PA-MPJPE on HCM and competitive PA-MPJPE on RICH. Additionally, PACE estimates much smoother motion by reducing the acceleration error (ACCEL) by 50\% on HCM and 56\% on RICH. 

\input{figures/cam_traj}
\vspace{1mm}
\noindent\textbf{Camera Motion Estimation.} The new HCM dataset provides ground-truth camera trajectories that allow us to benchmark camera motion estimation. Table~\ref{tab:hcm_ablation} shows that PACE substantially improves the camera motion estimated by a SOTA SLAM algorithm, \droid. Specifically, PACE reduces the camera translation error metric, ATE, by 12\% with scale alignment and 74\% without scale alignment. The above results show that our unified optimization approach can improve both human and camera motion estimation significantly, which answers the first two questions raised at the beginning of this section.

\noindent\textbf{Qualititative Comparison.} We also provide qualitative results to visualize the estimated human and camera motions in Fig.~\ref{fig:qualitative}. Please also refer to the \href{https://nvlabs.github.io/PACE/}{project page} for more qualitative results.

\noindent\textbf{Runtime.} It is worth noting that the runtime of our optimization framework increases \textit{sub-linearly} \wrt sequence length since we can optimize multiple chunks of the motion sequence simultaneously thanks to the parallel motion prior. On average, we can process sequences that are 1000 frames long in less than eight minutes. Notably, SLAHMR~\cite{ye2023slahmr} reports a runtime of 40 minutes for 100 frames, and this increases linearly with sequence length.

\subsection{Ablation Study}
We also conduct extensive ablation studies to investigate the effect of each optimization stage and important designs. As shown in Table~\ref{tab:hcm_ablation}, we compare the performance of PACE after each optimization stage: from stage-1 to stage-4, we gradually add variables to the global optimization -- camera trajectory transformation, global human orientation, local body pose, and full camera trajectory (see Sec.~\ref{sec:global_opt}) %
We observe that gradually adding additional variables to the optimization improves the human motion estimation results. %
We also try combining stages 2-4 and stages 3-4 to show the importance of multi-stage optimization. Combining these stages drops the performance compared to our 4-stage approach. %
We also compare PACE against the variant not using the point cloud loss (Stage-4 w/o $E_\textrm{PCL}$) in Table~\ref{tab:hcm_ablation}. We find that both human and camera motion estimation performance deteriorates when we do not use the point cloud loss, which shows that it is essential to use the background scene features in our unified optimization framework. During our experiments, we also evaluated the case when all variables are optimized from the beginning without stagewise optimization. We found that the optimization does not converge at all in this case.

%% file: tables/hcm_sota_slahmr.tex
\begin{table*}
    \centering
    \resizebox{\textwidth}{!}{
        \begin{tabular}{l|r|r|r|r|r|r|r|r}
        \toprule
        & \multicolumn{5}{c|}{\textbf{Human Motion Estimation}} & \multicolumn{3}{c}{\textbf{Camera Motion Estimation}} \\
        \textbf{Methods} & \textbf{\worldErr} $\downarrow$ & \textbf{\worldAlignedErr} $\downarrow$ & \textbf{\worldRootErr} $\downarrow$ & \textbf{\paErr} $\downarrow$ & \textbf{\accel} $\downarrow$ & \textbf{\ate} $\downarrow$ & \textbf{\ateWoS} $\downarrow$ & \textbf{\camAccel} $\downarrow$ \\
        \midrule 
        Initialization & 1116.3 & 650.0 & 1083.1 & 67.6 & 54.3 & 155.8 & 1670.7 & 17.1 \\
        \midrule
        Stage-1 (cam. traj. transform) & 1116.3 & 650.0 & 1083.1 & 67.6 & 54.3 & 155.8 & 643.0 & 17.1 \\
        Stage-2 (+ global human orientation) & 937.0 & 488.2 & 901.9 & 67.6 & 54.6 & 155.8 & 504.9 & 18.1 \\
        Stage-3 (+ local body pose) & 904.5 & 478.2 & 877.9 & 66.6 & 17.6 & 155.8 & 501.3 & 17.3 \\
        Stage-4 (+ full cam. traj.) w/o $E_\textrm{PCL}$ & 870.1 & 487.4 & 844.1 & 67.6 & 53.2 & 166.4 & 505.0 & \textbf{15.4} \\
        \midrule
        Stage 2-4 & 978.6 & 566.2 & 939.7 & 89.9 & 16.1 & 164.0 & 550.6 & 19.7 \\
        Stage 3-4 & 953.2 & 490.0 & 923.7 & 68.8 & 18.4 & 160.1 & 523.2 & 17.1 \\
        \midrule
        HybrIK~\cite{li2020hybrik} + SLAM~\cite{teed2021droid} & 1137.3 & 780.3 & 1100.9 & 67.6 & 51.3 & 155.8 & 1670.7 & 17.1 \\
        GLAMR~\cite{yuan2022glamr} & 1977.6 & 653.8 & 1958.0 & 86.0 & 33.4 & 1295.2 & 1714.6 & 282.9 \\
        SLAHMR~\cite{ye2023slahmr} & 888.9 & 483.5 & 862.2 & 69.9 & \textbf{14.9} & 155.8 & 506.5 & 17.6 \\
        \acronym (Ours) & \textbf{861.2} & \textbf{478.3} & \textbf{839.5} & \textbf{65.3} & 16.7 & \textbf{137.5} & \textbf{459.7} & 16.2 \\ 
        \bottomrule
        \end{tabular}
    }
   \vspace{-2mm}
   \caption{State-of-the-art comparison and ablation studies on the HCM dataset.}
   \vspace{-4mm}
   \label{tab:hcm_ablation}
\end{table*}

%% file: tables/rich_sota_slahmr.tex
\begin{table}
    \centering
    \resizebox{0.47\textwidth}{!}{

        \begin{tabular}{l|r|r|r|r|r}
        \toprule
        \textbf{Methods} & \worldErr $\downarrow$ & \\
        \worldAlignedErr $\downarrow$ & \worldRootErr $\downarrow$ & \paErr $\downarrow$ & \accel $\downarrow$ \\
        \midrule 
        HybrIK + SLAM & 1073.1 & 404.4 & 1066.2 & 46.7 & 20.2 \\
        GLAMR & 653.7 & 365.1 & 646.6 & 79.9 & 107.7 \\
        SLAHMR & 571.6 & 323.7 & 400.5 & 52.5 & 9.4 \\
        \acronym & \textbf{380.0} & \textbf{197.2} & \textbf{370.8} & \textbf{49.3} & \textbf{8.8} \\ 
        \bottomrule
        \end{tabular}
    }
   \vspace{-2mm}
   \caption{State of the art results on RICH dataset}
   \vspace{-6mm}
   \label{tab:rich_sota}
\end{table}

%% file: figures/hcm_dataset.tex
\begin{figure}  
\label{fig:hcm_dataset} 
\animategraphics[autoplay,width=1.0\linewidth]{25}{figures/hcm_dataset/00016/vis_}{00000}{00099} \\
\animategraphics[autoplay,width=1.0\linewidth]{25}{figures/hcm_dataset/00036/vis_}{00000}{00099} 
\vspace{-5mm}
\caption{Some examples of our proposed HCM dataset.  (animated figure, see in Adobe Acrobat).}
\vspace{-5mm}
\label{fig:hcm_dataset} 
\end{figure}

%% file: figures/qualitative.tex
\begin{figure*}[t]
    \centering
    \vspace{-5mm}
    \includegraphics[width=\linewidth]{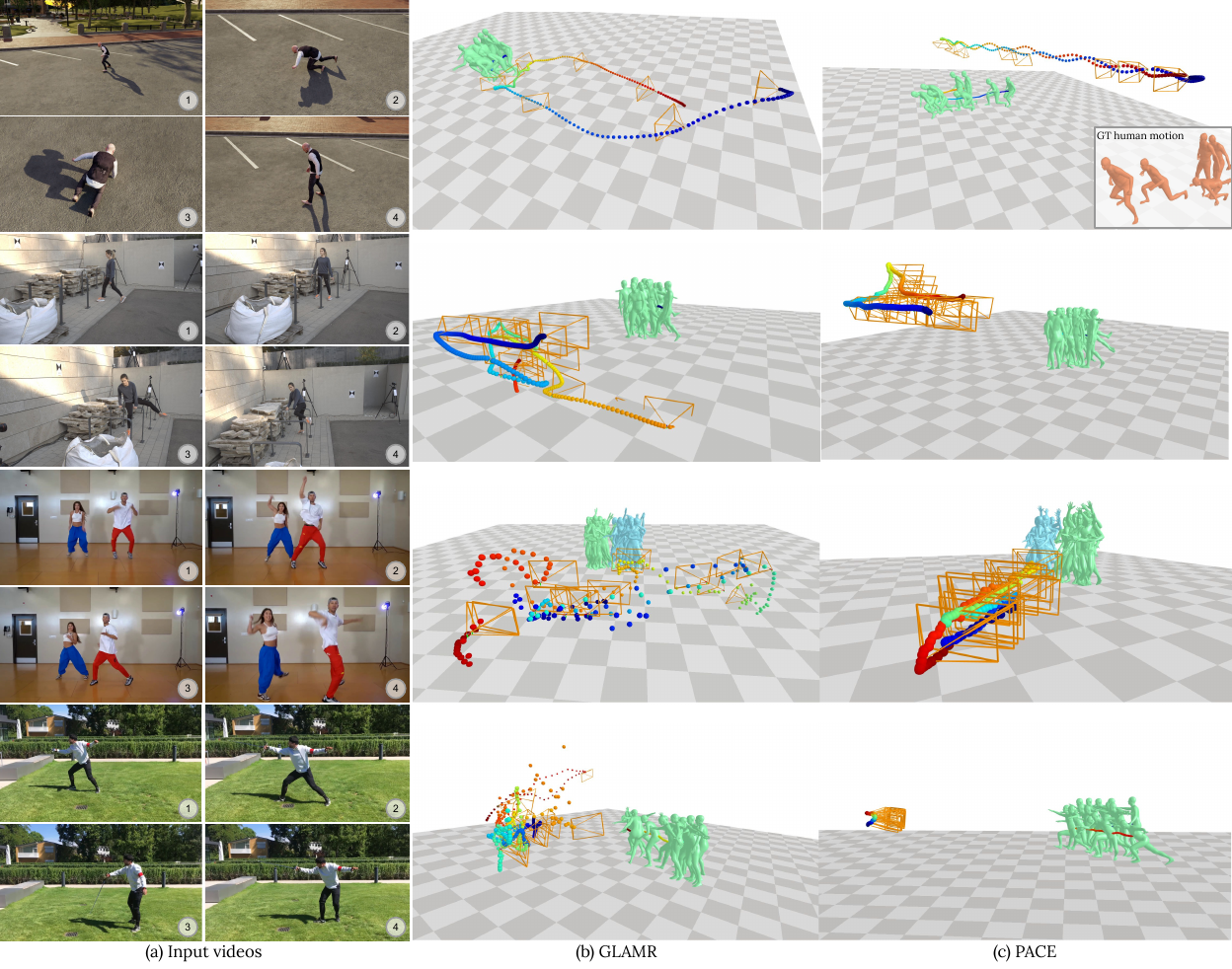}
    \vspace{-8mm}
    \caption{\textbf{Qualitative results} on HCM (row 1), RICH (row 2), and in-the-wild videos (rows 3 \& 4). PACE can estimate more accurate human and camera motion than the SOTA, GLAMR~\cite{yuan2022glamr}, for both datasets and in-the-wild videos.} 
    \label{fig:qualitative}
    \vspace{-5mm}
\end{figure*}{}

%% file: figures/cam_traj.tex
\begin{figure}[t]
    \centering
    \includegraphics[width=\linewidth]{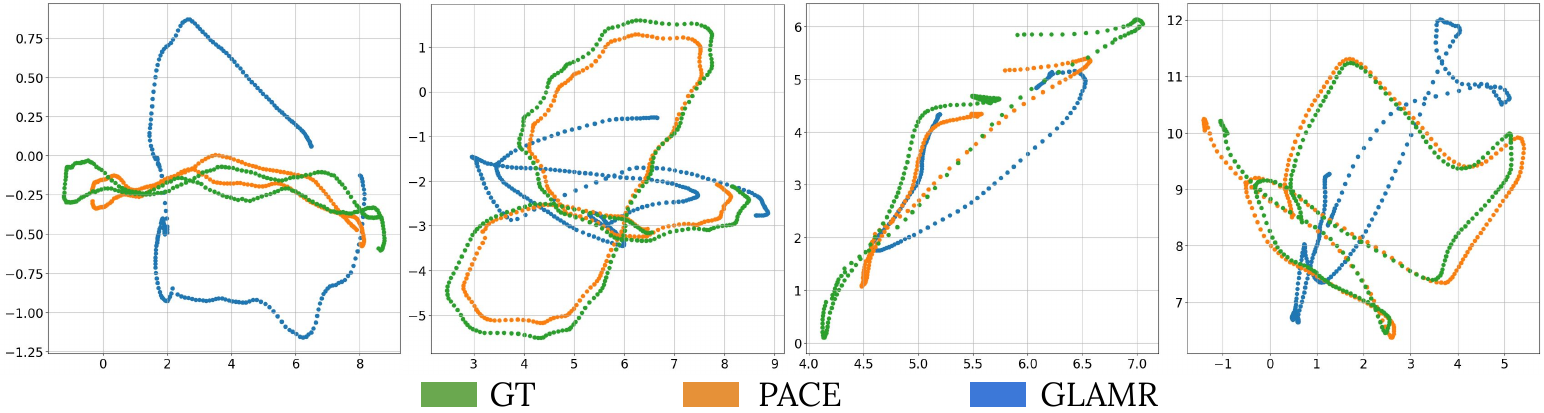}
    \vspace{-7mm}
    \caption{Comparison of camera motion estimation on HCM dataset. \acronym estimates more accurate camera motions compared to GLAMR.} 
    \label{fig:cam_traj}
    \vspace{-2ex}
\end{figure}{}

%% file: sections/05_conclusion.tex
\section{Conclusion}
We presented \acronym, a novel approach for accurate global human and camera motion estimation from dynamic cameras. Our approach leverages the complementary benefits of human motion priors and SLAM methods and integrates them into a unified optimization framework that jointly optimizes human and camera motions. We also introduced a new synthetic dataset called HCM for benchmarking global human and camera motion estimation. We demonstrated that our approach achieves superior performance as compared to the state-of-the-art methods in accurately recovering both human and camera motion.

Although our method can refine camera trajectories obtained from SLAM, it may not be effective in scenarios where SLAM methods fail catastrophically. We believe that the integration of physics-based constraints to prevent camera errors from overriding human motion priors would be an interesting future direction. %
Another limitation of our method is the assumption of a planar ground caused by the lack of scene annotation in the AMASS dataset. Also, while our proposed optimization is efficient, it is not real-time and requires batch processing to exploit future and past temporal information. Jointly solving camera and human motion in real-time and online fashion is
a significant challenge.

%% file: sections/06_appendix.tex
\section{Appendix}
In this appendix, we provide results on an additional dataset, EgoBody~\cite{Zhang:ECCV:2022}, and also provide additional implementation details. 

\subsection{Experiments on EgoBody dataset}
\input{tables/egobody_sota}

EgoBody~\cite{Zhang:ECCV:2022} is a large-scale dataset capturing ground-truth 3D human motions during social interactions in 3D scenes. EgoBody is captured with a head-mounted camera on an interactor, who sees and interacts with a second interactee. The camera moves as the interactor moves, and the ground truth 3D poses of the interactee are recorded in the world coordinate frame.  We follow~\cite{ye2023slahmr} and use the validation split of the dataset for evaluation. We use DROID-SLAM with the ground-truth camera intrinsics provided by the dataset. 

Table~\ref{tab:egobody_sota} compares \acronym with the state-of-the-art methods GLAMR~\cite{yuan2022glamr} and SLAHMR~\cite{ye2023slahmr}. As the results indicate, \acronym significantly outperforms GLAMR while achieving performance on par with SLAHMR in terms of accuracy. However, \acronym offers a significant computational advantage over SLAHMR, being up to 50 times faster for a sequence with 1000 frames. Note that the runtime of SLAHMR grows linearly with the sequence length, whereas our runtime increases sub-linearly. This improvement in efficiency demonstrates the potential of \acronym as a practical and effective solution for human and camera motion estimation from videos.

\subsection{Global optimization implementation details}
We empirically chose the weights of all error terms involved in the optimization, as summarized in Table~\ref{tab:opt_stages_detailed}. 

\input{tables/opt_stages_detailed.tex}

\subsection{HCM dataset generation}
\label{sec:hcm_dataset}
To create our HCM (Human and Camera Motion) dataset we used the characters from the RenderPeople~\cite{renderpeople} dataset with 3D scenes from the Unreal Engine Marketplace~\cite{unreal}. We manually labeled the navigable areas in each 3D scene \ie, sufficiently large, unobstructed flat areas within the scene. To generate a sequence, we randomly selected a 3D scene and a navigable area within it. We also randomly chose the number of people to be animated in the scene, ranging from 1 to 8 individuals. For each person, we selected a motion sequence from the validation set of the AMASS~\cite{AMASS:ICCV:2019} dataset. To ensure that each person's motion sequence was optimized for the scene, we iteratively added one person at a time. We optimized their global translation to ensure that they remained within the bounds of the navigable area and did not intersect with existing people in the scene. We also check the terrain height of the navigable area and adjusted each character's root translation accordingly to ensure they were at the correct height relative to the terrain. Finally, we rendered the animated 3D scene into a video sequence using a moving camera. To generate camera trajectories, we designed heuristics
to replicate typical camera movements observed in everyday videos and professional movies. More specifically, we used dolly zoom, random arc motion towards a person, camera motions from the MannequinChallenge dataset~\cite{li2019mannequin}, cameras tracking a specific person, etc. This approach allowed us to generate a diverse set of sequences with varying numbers of people and diverse body and camera motions. In total, we generated 25 video sequences for evaluation. Some examples can be seen in the \href{https://nvlabs.github.io/PACE/}{project page}. We believe our HCM dataset will be extremely useful for evaluating human and camera motion estimation methods and furthering research in this direction. 




%% file: tables/egobody_sota.tex
\begin{table}[t]
    \centering
    \resizebox{\linewidth}{!}{

        \begin{tabular}{l|r|r|r|r|r}
        \toprule
        \textbf{Methods} & \worldErr $\downarrow$ & \worldAlignedErr$\downarrow$ & \paErr $\downarrow$ & \accel $\downarrow$ & Runtime (per 1000 imgs)\\
        \midrule 
        GLAMR~\cite{yuan2022glamr} & 416.1 & 239.0 & 114.3 & 173.5 & 7min \\
        SLAHMR~\cite{ye2023slahmr} & \textbf{141.1} & 101.2 & 79.13 & 25.78 & 400min \\
        \acronym (Ours) & 147.9 & \textbf{101.0} & \textbf{66.5} & \textbf{6.7} & 8min \\
        \bottomrule
        \end{tabular}
    }
   \caption{State-of-the-art results on the EgoBody dataset.}
   \label{tab:egobody_sota}
\end{table}

%% file: tables/opt_stages_detailed.tex
\begin{table*}[tb]
    \centering
    \resizebox{\linewidth}{!}{

        \begin{tabular}{l|c|c|c}
        \toprule
        \textbf{Stages} & \textbf{Opt. Variables} & \textbf{Error Functions} & \textbf{Learning rate (lr) \& Weights}\\
        \midrule 
        Stage-1 & $s, h_0, R_0, \beta$ & $E_{2D}+E_\beta$ & $\textrm{lr}=0.01, \lambda_{2D}=0.001, \lambda_{\beta}=1.0 $ \\
        \midrule 
        Stage-2 & $s, h_0, R_0, \beta, \mathbf{z}_\Phi$ & $E_\textrm{body}+E_\textrm{scene}$ & $\textrm{lr}=0.01, \lambda_\textrm{2D}=0.001, \lambda_{\beta}=1, \lambda_\textrm{contact}=100, \lambda_\textrm{height}=10, \lambda_\textrm{VAE}=0.1, \lambda_\textrm{consist}=1, \lambda_\textrm{smooth}=1$\\
        \midrule 
        Stage-3 & $s, h_0, R_0, \beta, \mathbf{z}_\Phi, \mathbf{z}_\phi$ & $E_\textrm{body}+E_\textrm{scene}$ & $\textrm{lr}=0.01, \lambda_\textrm{2D}=0.001, \lambda_{\beta}=1, \lambda_\textrm{contact}=100, \lambda_\textrm{height}=10, \lambda_\textrm{VAE}=0.1, \lambda_\textrm{consist}=1, \lambda_\textrm{smooth}=1, \lambda_\textrm{pose}=1$\\
        \midrule 
        Stage-4 & $\beta, \mathbf{z}_\Phi, \mathbf{z}_\phi, R_t, T_t$ & $E_\textrm{body}+E_\textrm{scene}+E_\textrm{camera}$ & $\textrm{lr}=0.001, \lambda_\textrm{2D}=0.001, \lambda_{\beta}=1, \lambda_\textrm{contact}=100, \lambda_\textrm{height}=10, \lambda_\textrm{VAE}=0.1, \lambda_\textrm{consist}=1, \lambda_\textrm{smooth}=1, \lambda_\textrm{pose}=1, \lambda_\textrm{PCL}=1e^{-4}$\\
        \bottomrule
        \end{tabular} 
    }
   \caption{Optimization stages and weights. 
   }
    \label{tab:opt_stages_detailed}
\end{table*}